# Regularized Robust Coding for Face Recognition

Meng Yang[a], *Student Member, IEEE,* Lei Zhang[a,1], *Member, IEEE*
Jian Yang[b], *Member, IEEE,* and David Zhang[a], *Fellow, IEEE*

[a]Dept. of Computing, The Hong Kong Polytechnic University, Hong Kong, China
[b]School of Computer Science and Technology, Nanjing Univ. of Science and Technology, Nanjing, China

**Abstract:** Recently the sparse representation based classification (SRC) has been proposed for robust face recognition (FR). In SRC, the testing image is coded as a sparse linear combination of the training samples, and the representation fidelity is measured by the $l_2$-norm or $l_1$-norm of the coding residual. Such a sparse coding model assumes that the coding residual follows Gaussian or Laplacian distribution, which may not be effective enough to describe the coding residual in practical FR systems. Meanwhile, the sparsity constraint on the coding coefficients makes SRC's computational cost very high. In this paper, we propose a new face coding model, namely regularized robust coding (RRC), which could robustly regress a given signal with regularized regression coefficients. By assuming that the coding residual and the coding coefficient are respectively independent and identically distributed, the RRC seeks for a maximum a posterior solution of the coding problem. An iteratively reweighted regularized robust coding (IR$^3$C) algorithm is proposed to solve the RRC model efficiently. Extensive experiments on representative face databases demonstrate that the RRC is much more effective and efficient than state-of-the-art sparse representation based methods in dealing with face occlusion, corruption, lighting and expression changes, etc.

*Keywords:* Robust coding, face recognition, sparse representation, regularization

___________________

[1] Corresponding author. Email: cslzhang@comp.polyu.edu.hk. This work is supported by the Hong Kong RGC General Research Fund (PolyU 5351/08E).

# 1. Introduction

As one of the most visible and challenging problems in computer vision and pattern recognition, face recognition (FR) has been extensively studied in the past two decades [1-15][18-20][54-57], and many representative methods, such as Eigenface [2], Fisherface [3] and SVM [4], have been proposed. Moreover, to deal with the challenges in practical FR system, active shape model and active appearance model [5] were developed for face alignment; LBP [6] and its variants were used to deal with illumination changes; and Eigenimages [7-8] and probabilistic local approach [9] were proposed for FR with occlusion. Although much progress have been made, robust FR to occlusion/corruption is still a challenging issue because of the variations of occlusion, such as disguise, continuous or pixel-wise occlusion, randomness of occlusion position and the intensity of occluded pixels.

The recognition of a query face image is usually accomplished by classifying the features extracted from this image. The most popular classifier for FR may be the nearest neighbor (NN) classifier due to its simplicity and efficiency. In order to overcome NN's limitation that only one training sample is used to represent the query face image, Li and Lu proposed the nearest feature line (NFL) classifier [10], which uses two training samples for each class to represent the query face. Chien and Wu [11] then proposed the nearest feature plane (NSP) classifier, which uses three samples to represent the test image. Later on, classifiers using more training samples for face representation were proposed, such as the local subspace classifier (LSC) [12] and the nearest subspace (NS) classifiers [11, 13-15], which represent the query sample by all the training samples of each class. Though NFL, NSP, LSC and NS achieve better performance than NN, all these methods with holistic face features are not robust to face occlusion.

Generally speaking, these nearest classifiers, including NN, NFL, NFP, LSC and NS, aim to find a suitable representation of the query face image, and classify it by checking which class can give a better representation than other classes. Nonetheless, how to formulate the representation model for classification tasks such as FR is still a challenging problem. In recent years, sparse representation (or sparse coding) has been attracting a lot of attention due to its great success in image processing [16, 17], and it has also been used for FR [18, 19, 20] and texture classification [21, 22]. Based on the findings that natural images can be generally coded by structural primitives (e.g., edges and line segments) that are qualitatively similar in form to simple cell receptive fields [23], sparse coding represents a signal using a small number of atoms parsimoniously chosen out of an over-complete dictionary. The sparsity of the coding coefficient can be measured by $l_0$-norm, which counts the number of nonzero entries in a vector. Since the combinatorial $l_0$-norm minimization is an NP-hard problem, the $l_1$-norm minimization, as the closest convex function to $l_0$-norm minimization, is widely employed in sparse coding, and it has been shown that $l_0$-norm and $l_1$-norm minimizations are equivalent if the solution is sufficiently sparse [24]. In general, the sparse coding problem can be formulated as

$$\min_{\alpha} \|\alpha\|_1 \quad \text{s.t.} \quad \|y - D\alpha\|_2^2 \leq \varepsilon \qquad (1)$$

where $y$ is the given signal, $D$ is the dictionary of coding atoms, $\alpha$ is the coding vector of $y$ over $D$, and $\varepsilon > 0$ is a constant. Recently, Wright *et al.* [18] applied sparse coding to FR and proposed the sparse representation based classification (SRC) scheme. By coding a query image $y$ as a sparse linear combination of all the training samples via Eq. (1), SRC classifies $y$ by evaluating which class could result in the minimal reconstruction error of it. However, it has been indicated in [25] that the success of SRC actually owes to its utilization of collaborative representation on the query image but not the $l_1$-norm sparsity constraint on coding coefficient.

One interesting feature of SRC is its processing of face occlusion and corruption. More specifically, it introduces an identity matrix $I$ as a dictionary to code the outlier pixels (e.g., corrupted or occluded pixels):

$$\min_{\alpha} \|[\alpha; \beta]\|_1 \quad \text{s.t.} \quad y = [D, I] \cdot [\alpha; \beta] \qquad (2)$$

By solving Eq. (2), SRC shows good robustness to face occlusions such as block occlusion, pixel corruption and disguise. It is not difficult to see that Eq. (2) is basically equivalent to $\min_{\alpha} \|\alpha\|_1$ s.t. $\|y - D\alpha\|_1 < \varepsilon$. That is, it uses $l_1$-norm to model the coding residual $y$-$D\alpha$ to gain certain robustness to outliers.

The SRC has close relationship to the nearest classifiers. Like NN, NFL [10], NFP [11], LSC [12] and NS classifiers [13-15, 11], SRC also represents the query sample as the linear combination of training samples; however, it forces the representation coefficients being sparse (instead of presetting the number of non-zero representation coefficients) and allows across-class representation (i.e., significant coding coefficients can be from samples of different classes). SRC could be seen as a more general model than the previous nearest classifiers, and it uses the samples from all classes to collaboratively represent the query sample to overcome the small-sample-size problem in FR. In addition, different from the methods such as [6, 9, 26,58] which use local region features, color features or gradient information to handle some special occlusion , SRC shows interesting results in dealing with occlusion by assuming a sparse coding residual, as in Eq. (2). There are many following works to extend and improve SRC, such as feature-based SRC [20], SRC for face misalignment or pose variation [27-28], and SRC for continuous occlusion [29].

Although the sparse coding model in Eq. (1) has made a great success in image restoration [16-17] and led to interesting results in FR [18-20], there are two issues to be considered more carefully when applying it to pattern classification tasks such as FR. One is that whether the $l_1$-sparsity constraint $\|\ \|_1$ is indispensable to regularize the solution, since the $l_1$-minimization needs much computational cost. The other is that whether the term $\|y - D\alpha\|_2^2 \leq \varepsilon$ is effective enough to characterize the signal fidelity, especially when the observation $y$ is noisy and/or has many outliers. For the first issue, on one side reweighted $l_1$ or $l_2$ minimization was proposed to speed up the sparse coding process [30, 52]; one the other side some works [25, 31, 49] have questioned the use of sparse coding for image classification. Particularly, Zhang *et al.* [25] have shown that it is not necessary to impose the $l_1$-sparsity constraint on the coding vector $\alpha$, while the $l_2$-norm regularization on $\alpha$ performs equally well. Zhang *et al*. also indicated that the success of SRC actually comes from its collaborative representation of $y$ over all classes of training samples. For the second issue, to the best of our knowledge, few works have been reported in the scheme of sparse representation except for the $l_1$-norm fidelity (i.e., $\|y - D\alpha\|_1 \leq \varepsilon$) in [18-19], the correntropy based Gaussian-kernel fidelity in [32-33] and our previous work in [35]. The fidelity term has a very high impact on the final coding result. From the viewpoint of *maximum a posterior* (MAP) estimation, defining the fidelity term with $l_2$- or $l_1$-norm actually assumes that the coding residual $e=y-D\alpha$ follows Gaussian or Laplacian distribution. In practice, however, such an assumption may not hold well, especially when occlusions, corruptions and expression variations occur in the query face images. Although Gaussian kernel based fidelity term utilized in [32-33] is claimed to be robust to non-Gaussian noise [34], it may not work well in FR with occlusion due to the complex variation of occlusion. For example, the scarf disguise occlusion needs to be manually removed in [33].

To increase the robustness of FR to occlusion, pixel corruption, disguises and big expression variations, etc., we propose a regularized robust coding (RRC) model in this paper. A special case of RRC, namely robust sparse coding (RSC), has been presented in our previous work [35] by assuming that the coding coefficients are sparse. Although RSC achieves state-of-the-art FR results, the $l_1$-sparsity constraint on the coding vector $\alpha$ makes the computational cost very high. In this paper, we assume that the coding residual $e$ and the coding vector $\alpha$ are respectively independent and identically distributed, and then robustly regress the given signal based on the MAP principle. In implementation, the RRC minimization problem is transformed into an iteratively reweighted regularized robust coding (IR$^3$C) problem with a reasonably designed weight function for robust FR. Our extensive experiments in benchmark face databases show that RRC achieves much better performance than existing sparse representation based FR methods, especially when there are complicated variations, such as face occlusions, corruptions and expression changes, etc.

The rest of this paper is organized as follows. Section 2 presents the proposed RRC model. Section 3 presents the algorithm of RRC. Section 4 conducts the experiments, and Section 5 concludes the paper.

## 2. Regularized Robust Coding (RRC)

### 2.1. The modeling of RRC

The conventional sparse coding model in Eq. (1) is equivalent to the so-called LASSO problem [38]:

$$\min_{\alpha} \|y - D\alpha\|_2^2 \quad \text{s.t.} \quad \|\alpha\|_1 \leq \sigma \tag{3}$$

where $\sigma > 0$ is a constant, $y=[y_1;y_2;\ldots;y_n] \in \Re^n$ is the signal to be coded, $D=[d_1, d_2, \ldots, d_m] \in \Re^{n \times m}$ is the dictionary with column vector $d_j$ being its $j^{\text{th}}$ atom, and $\alpha \in \Re^m$ is the vector of coding coefficients. In the problem of face recognition (FR), the atom $d_j$ can be simply set as the training face sample (or its dimensionality reduced feature) and hence the dictionary $D$ can be the whole training dataset.

If we have the prior that the coding residual $e = y - D\alpha$ follows Gaussian distribution, the solution to Eq. (3) will be the maximum likelihood estimation (MLE) solution. If $e$ follows Laplacian distribution, the $l_1$-sparsity constrained MLE solution will be

$$\min_{\alpha} \|y - D\alpha\|_1 \quad \text{s.t.} \quad \|\alpha\|_1 \leq \sigma \tag{4}$$

The above Eq. (4) is essentially another expression of Eq. (2) because they have the same Lagrangian formulation: $\min_{\alpha}\{\|y-D\alpha\|_1+\lambda\|\alpha\|_1\}$ [39].

In practice, however, the Gaussian or Laplacian priors on $e$ may be invalid, especially when the face image $y$ is occluded, corrupted, etc. Let's use examples to illustrate the fitted distributions of residual $e$ by different models. Fig. 1(a) shows a clean face image, denoted by $y_o$, while Fig. 1(b) and Fig. 1(c) show the occluded and corrupted query images $y$, respectively. The residual is computed as $e = y - D\hat{\alpha}$, while to make the coding vector more accurate we use the clean image to calculate it via Eq. (3): $\hat{\alpha} = \arg\min_{\alpha} \|y_o - D\alpha\|_2^2 \quad \text{s.t.} \quad \|\alpha\|_1 \leq \sigma$. The empirical and fitted distributions of *e* by using Gaussian, Laplacian and the distribution model (refer to Eq. (15)) associated with the proposed method are plotted in Fig. 1(d). Fig. 1(e) shows the distributions in log domain for better observation of the tails. It can be seen that the empirical distribution of *e* has a strong peak at zero but a long tail, which is mostly caused by the occluded and corrupted pixels. For robust FR, a good fitting of the tail is much more important than the fitting of the peak, which is produced by the small trivial coding errors. It can be seen from Fig. 1(e) that the proposed model can well fit the heavy tail of the empirical distribution, much better than the Gaussian and Laplacian models. Meanwhile, Laplacian works better than Gaussian in fitting the heavy tail, which explains why the sparse coding model in Eq. (4) (or Eq. (2)) works better than the model in Eq. (1) (or Eq. (3)) in handling face occlusion and corruption.

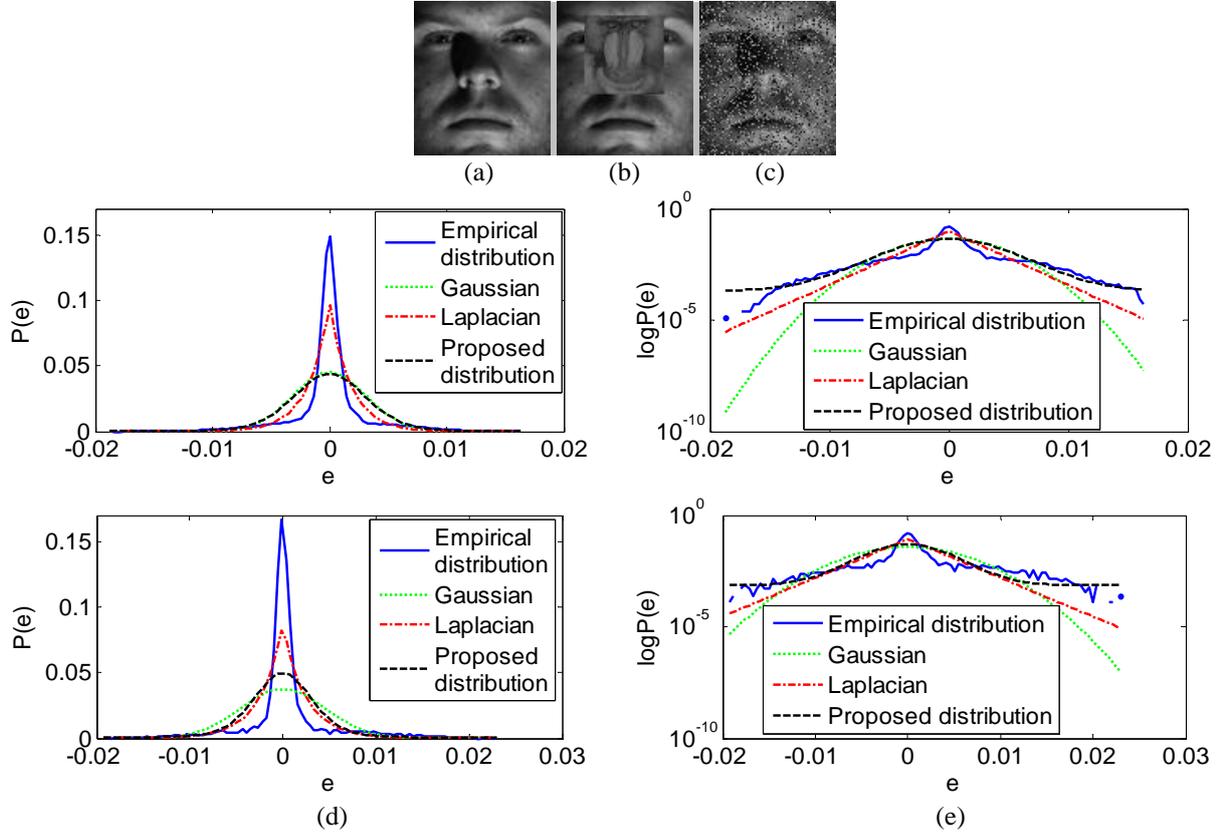

**Figure 1:** The empirical distribution of coding residuals and the fitted distributions by different models. (a) Clean face image; (b) and (c) are occluded and corrupted query face images; (d) and (e) show the distributions (top row: occluded image; bottom row: corrupted image) of coding residuals in linear and log domains, respectively.

Inspired by the robust regression theory [36][37][53], in our previous work [35] we proposed an MLE solution for robust face image representation. Rewrite $D$ as $D = [r_1; r_2; \ldots; r_n]$, where $r_i$ is the $i^{th}$ row of $D$, and let $e = y - D\alpha = [e_1; e_2; \ldots; e_n]$, where $e_i = y_i - r_i\alpha$, $i=1,2,\ldots,n$. Assume that $e_1, e_2, \ldots, e_n$ are independent and identically distributed (i.i.d.) and the probability density function (PDF) of $e_i$ is $f_\theta(e_i)$, where $\theta$ denotes the unknown parameter set that characterizes the distribution, the so-called robust sparse coding (RSC) [35] was formulated as the following $l_1$-sparsity constrained MLE problem (let $\rho_\theta(e) = -\ln f_\theta(e)$)

$$\min_{\boldsymbol{\alpha}} \sum_{i=1}^{n} \rho_\theta (y_i - r_i\boldsymbol{\alpha}) \text{ s.t. } \|\boldsymbol{\alpha}\|_1 \leq \sigma \tag{5}$$

Like SRC, the above RSC model assumes that the coding coefficients are sparse and uses $l_1$-norm to characterize the sparsity. However, the $l_1$-sparsity constraint makes the complexity of RSC high, and recently it has been indicated in [25] that the $l_1$-sparsity constraint on $\boldsymbol{\alpha}$ is not the key for the success of SRC [18]. In this paper, we propose a more general model, namely regularized robust coding (RRC). The RRC can be much more efficient than RSC, while RSC is one specific instantiation of the RRC model.

Let's consider the face representation problem from a viewpoint of Bayesian estimation, more specifically, the *maximum a posterior* (MAP) estimation. By coding the query image *y* over a given a dictionary ***D***, the MAP estimation of the coding vector $\boldsymbol{\alpha}$ is $\hat{\boldsymbol{\alpha}} = \arg\max_{\boldsymbol{\alpha}} \ln P(\boldsymbol{\alpha} | y)$. Using the Bayesian formula, we have

$$\hat{\boldsymbol{\alpha}} = \arg\max_{\boldsymbol{\alpha}} \{\ln P(y|\boldsymbol{\alpha}) + \ln P(\boldsymbol{\alpha})\} \tag{6}$$

Assuming that the elements $e_i$ of coding residual $e=y-D\boldsymbol{\alpha} = [e_1; e_2; ...; e_n]$ are i.i.d. with PDF $f_\theta(e_i)$, we have $P(y|\boldsymbol{\alpha}) = \prod_{i=1}^{n} f_\theta(y_i - r_i\boldsymbol{\alpha})$. Meanwhile, assume that the elements $\alpha_j, j=1,2,...,m$, of the coding vector $\boldsymbol{\alpha}=[\alpha_1; \alpha_2; ...; \alpha_m]$ are i.i.d. with PDF $f_o(\alpha_j)$, there is $P(\boldsymbol{\alpha}) = \prod_{j=1}^{m} f_o(\alpha_j)$. The MAP estimation of $\boldsymbol{\alpha}$ in Eq. (6) is

$$\hat{\boldsymbol{\alpha}} = \arg\max_{\boldsymbol{\alpha}} \left\{ \prod_{i=1}^{n} f_\theta(y_i - r_i\boldsymbol{\alpha}) + \prod_{j=1}^{m} f_o(\alpha_j) \right\} \tag{7}$$

Letting $\rho_\theta(e) = -\ln f_\theta(e)$ and $\rho_o(\alpha) = -\ln f_o(\alpha)$, Eq. (7) is converted into

$$\hat{\boldsymbol{\alpha}} = \arg\min_{\boldsymbol{\alpha}} \left\{ \sum_{i=1}^{n} \rho_\theta(y_i - r_i\boldsymbol{\alpha}) + \sum_{j=1}^{m} \rho_o(\alpha_j) \right\} \tag{8}$$

We call the above model regularized robust coding (RRC) because the fidelity term $\sum_{i=1}^{n} \rho_\theta (y_i - r_i\boldsymbol{\alpha})$ will be very robust to outliers, while $\sum_{j=1}^{m} \rho_o(\alpha_j)$ is the regularization term depending on the prior probability $P(\boldsymbol{\alpha})$.

It can be seen that $\sum_{j=1}^{m} \rho_o(\alpha_j)$ becomes the $l_1$-norm sparse constraint when $\alpha_j$ is Laplacian distributed, i.e., $P(\boldsymbol{\alpha}) = \prod_{j=1}^{m} \exp(-\|\alpha_j\|_1/\sigma_\alpha)/2\sigma_\alpha$. For the problem of classification, it is desired that only the representation coefficients associated with the dictionary atoms from the target class could have big absolute values. As we do not know beforehand which class the query image belongs to, a reasonable prior can be that only a small percent of representation coefficients have significant values. Therefore, we assume that the representation coefficient $\alpha_j$ follows generalized Gaussian distribution (GGD). There is

$$f_o(\alpha_j) = \beta \exp\left\{-\left(|\alpha_j|/\sigma_\alpha\right)^\beta\right\} \Big/ \left(2\sigma_\alpha \Gamma(1/\beta)\right) \tag{9}$$

where $\Gamma$ denotes the gamma function.

For the representation residual, it is difficult to predefine the distribution due to the diversity of image variations. In general, we assume that the unknown PDF $f_\theta(e)$ are symmetric, differentiable, and monotonic w.r.t. $|e|$, respectively. So $\rho_\theta(e)$ has the following properties: (1) $\rho_\theta(0)$ is the global minimal of $\rho_\theta(x)$; (2) symmetry: $\rho_\theta(x) = \rho_\theta(-x)$; (3) monotonicity: $\rho_\theta(x_1) > \rho_\theta(x_2)$ if $|x_1| > |x_2|$. Without loss of generality, we let $\rho_\theta(0)=0$.

Two key issues in solving the RRC model are how to determine the distributions $\rho_\theta$ (or $f_\theta$), and how to minimize the energy functional. Simply taking $f_\theta$ as Gaussian or Laplacian and taking $f_o$ as Laplacian, the RRC model will degenerate to the conventional sparse coding problem in Eq. (3) or Eq. (4). However, as we showed in Fig. 1, such preset distributions for $f_\theta$ have much bias and are not robust enough to outliers, and the Laplacian setting of $f_o$ makes the minimization inefficient. In this paper, we allow $f_\theta$ to have a more flexible shape, which is adaptive to the input query image $y$ so that the system is more robust to outliers. To this end, we transform the minimization of Eq. (8) into an iteratively reweighted regularized coding problem in order to obtain the approximated MAP solution of RRC effectively and efficiently.

## 2.2. RRC via iteratively reweighting

Let $F_\theta(e) = \sum_{i=1}^{n} \rho_\theta(e_i)$. The Taylor expansion of $F_\theta(e)$ in the neighborhood of $e_0$ is:

$$\tilde{F}_\theta(e) = F_\theta(e_0) + (e-e_0)^T F'_\theta(e_0) + R_1(e) \tag{10}$$

where $R_1(e)$ is the high order residual, and $F'_\theta(e)$ is the derivative of $F_\theta(e)$. Denote by $\rho'_\theta$ the derivative of $\rho_\theta$, and there is $F'_\theta(e_0) = \left[\rho'_\theta(e_{0,1}); \rho'_\theta(e_{0,2}); \cdots; \rho'_\theta(e_{0,n})\right]$, where $e_{0,i}$ is the $i^{th}$ element of $e_0$. To make $F'_\theta(e)$ strictly convex for easier minimization, we approximate the residual term as $R_1(e) \approx 0.5(e-e_0)^T W(e-e_0)$, where $W$ is a diagonal matrix for that the elements in $e$ are independent and there is no cross term of $e_i$ and $e_j$, $i \neq j$, in $F_\theta(e)$.

Since $F_\theta(e)$ reaches its minimal value (i.e., 0) at $e=0$, we also require that its approximation $\tilde{F}_\theta(e)$ reaches the minimum at $e=0$. Letting $\tilde{F}'_\theta(0) = 0$, we have the diagonal elements of $W$ as

$$W_{i,i} = \omega_{\theta}(e_{0,i}) = \rho'_{\theta}(e_{0,i})/e_{0,i} \qquad (11)$$

According to the properties of $\rho_\theta$, we know that $\rho'_\theta(e_i)$ will have the same sign as $e_i$. So $W_{i,i}$ is a non-negative scalar. Then $\tilde{F}_\theta(e)$ can be written as

$$\tilde{F}_\theta(e) = \frac{1}{2}\|W^{1/2}e\|_2^2 + b_{e_0} \qquad (12)$$

where $b_{e_0} = \sum_{i=1}^{n}\left[\rho_\theta(e_{0,i}) - \rho'_\theta(e_{0,i})e_{0,i}/2\right]$ is a scalar constant determined by $e_0$.

Without considering the constant $b_{e_0}$, the RRC model in Eq. (8) could be approximated as

$$\hat{\alpha} = \arg\min_{\alpha} \left\{ \frac{1}{2}\|W^{1/2}(y - D\alpha)\|_2^2 + \sum_{j=1}^{m}\rho_o(\alpha_j) \right\} \qquad (13)$$

Certainly, Eq. (13) is a local approximation of Eq. (8) but it makes the minimization of RRC feasible via iteratively reweighted $l_2$-regularized coding, in which $W$ is updated via Eq. (11). Now, the minimization of RRC is turned to how to calculate the diagonal weight matrix $W$.

### 2.3. The weights $W$

The element $W_{i,i}$, i.e., $\omega_\theta(e_i)$, is the weight assigned to pixel $i$ of query image $y$. Intuitively, in FR the outlier pixels (e.g., occluded or corrupted pixels) should have small weights to reduce their effect on coding $y$ over $D$. Since the dictionary $D$, composed of non-occluded/non-corrupted training face images, could well represent the facial parts, the outlier pixels will have rather big coding residuals. Thus, the pixel which has a big residual $e_i$ should have a small weight. Such a principle can be observed from Eq. (11), where $\omega_\theta(e_i)$ is inversely proportional to $e_i$ and modulated by $\rho'_\theta(e_i)$. Refer to Eq. (11), since $\rho_\theta$ is differentiable, symmetric, monotonic and has its minimum at origin, we can assume that $\omega_\theta(e_i)$ is continuous and symmetric, while being inversely proportional to $e_i$ but bounded (to increase stability). Without loss of generality, we let $\omega_\theta(e_i) \in [0, 1]$. With these considerations, one good choice of $\omega_\theta(e_i)$ is the widely used logistic function [40]:

$$\omega_\theta(e_i) = \exp(-\mu e_i^2 + \mu\delta)/(1 + \exp(-\mu e_i^2 + \mu\delta)) \qquad (14)$$

where $\mu$ and $\delta$ are positive scalars. Parameter $\mu$ controls the decreasing rate from 1 to 0, and $\delta$ controls the location of demarcation point. Here the value of $\mu\delta$ should be big enough to make $\omega_\theta(0)$ close to 1 (usually we set $\mu\delta \geqslant 8$). With Eq. (14), Eq. (11) and $\rho_\theta(0)=0$, we could get

$$\rho_\theta(e_i) = -\frac{1}{2\mu}\left(\ln\left(1+\exp(-\mu e_i^2 + \mu\delta)\right) - \ln(1+\exp\mu\delta)\right) \tag{15}$$

We can see that the above $\rho_\theta$ satisfies all the assumptions and properties discussed in Section 2.1.

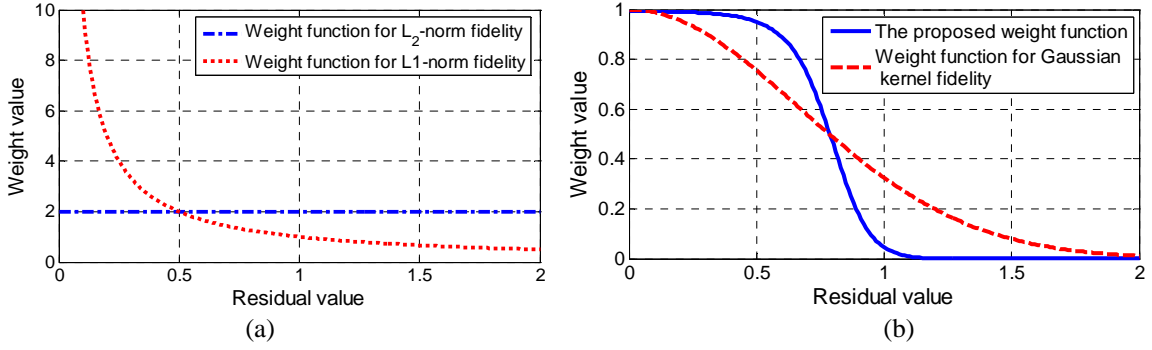

**Figure 2:** Weight functions for different signal fidelity terms, including (a) $l_2$ and $l_1$-norm fidelity terms in SRC [18] and (b) the Gaussian kernel fidelity term [32][33], as well as the proposed RRC fidelity term.

The PDF $f_\theta$ associated with $\rho_\theta$ in Eq.(15) is more flexible than the Gaussian and Laplacian functions to model the residual $e$. It can have a longer tail to address the residuals yielded by outlier pixels such as corruptions and occlusions (refer to Fig. 1 for examples), and hence the coding vector $\alpha$ will be robust to the outliers in $y$. $\omega_\theta(e_i)$ could also be set as other functions. However, as indicated by [59], the proposed logistic weight function is the binary classifier derived via MAP estimation, which is suitable to distinguish inliers and outliers. When $\omega_\theta(e_i)$ is set as a constant such as $\omega_\theta(e_i)=2$, it corresponds to the $l_2$-norm fidelity in Eq. (3); when set as $\omega_\theta(e_i)=1/|e_i|$, it corresponds to the $l_1$-norm fidelity in Eq. (4); when set as a Gaussian function $\omega_\theta(e_i) = \exp(-e_i^2/2\sigma^2)$, it corresponds to the Gaussian kernel fidelity in [32-33]. However, all these functions are not as robust as Eq. (14) to outliers, as illustrated in Fig. 2. From Fig. 2, one can see that the $l_2$-norm fidelity treats all pixels equally, no matter it is outlier or not; the $l_1$-norm fidelity assigns higher weights to pixels with smaller residuals; however, the weight can be infinity when the residual approaches to zero, making the coding unstable. Both our proposed weight function and the weight function of the Gaussian fidelity used in [32-33] are bounded in [0, 1], and they have an intersection point with weight value as 0.5. However, the proposed weight function prefers to assign larger weights to inliers and smaller weights to outliers; that is, it has higher capability to classify inliers and outliers.

The sparse coding models in Eqs. (3) and (4) are instantiations of the RRC model in Eq.(13) with $\beta=1$ in Eq.(9). The model in Eq. (3) is the case by letting $\omega_\theta(e_i)=2$. The model in Eq. (4) is the case by letting $\omega_\theta(e_i)=1/|e_i|$. Compared with the models in Eqs. (3) and (4), the proposed RRC model (Eq. (8) or Eq. (13)) is much more robust to outliers (usually the pixels with big residuals) because it will adaptively assign small weights to them. Although the model in Eq. (4) also assigns small weights to outliers, its weight function $\omega_\theta(e_i)=1/|e_i|$ is not bounded (i.e., the weights assigned to very small residuals can have very big values and dramatic changing ratios), making it less effective to distinguish between inliers and outliers.

### 2.4. Two important cases of RRC

The minimization of RRC model in Eq. (13) can be accomplished iteratively, while in each iteration $W$ and $\alpha$ are updated alternatively. By fixing the weight matrix $W$, the RRC with GGD prior on representation (i.e., Eq. (9)) and $\rho_o(\alpha) = -\ln f_o(\alpha)$ could be written as

$$\hat{\alpha} = \arg\min_\alpha \left\{ \frac{1}{2} \left\| W^{1/2}(y - D\alpha) \right\|_2^2 + \sum_{j=1}^m \left( \lambda |\alpha_j|^\beta + b_{\alpha_0} \right) \right\} \tag{16}$$

where $\rho_o(\alpha_j) = \lambda |\alpha_j|^\beta + b_{\alpha_0}$, $\lambda = (1/\sigma_\alpha)^\beta$ and $b_{\alpha_0} = \ln(2\sigma_\alpha \Gamma(1/\beta)/\beta)$ is a constant. Similar to the processing of $F_\theta(e) = \sum_{i=1}^n \rho_\theta(e_i)$ in Section 2.2, $\sum_{j=1}^m \rho_o(\alpha_j)$ could also be approximated by the Taylor expansion. Then Eq. (16) changes to

$$\hat{\alpha} = \arg\min_\alpha \left\{ \left\| W^{1/2}(y - D\alpha) \right\|_2^2 + \sum_{j=1}^m V_{j,j} \alpha_j^2 \right\} \tag{17}$$

where $W$ is a diagonal matrix with $V_{j,j} = \rho_o'(\alpha_j)/\alpha_j$.

The value of $\beta$ determines the types of regularization. If $0<\beta\leq 1$, then sparse regularization is applied; otherwise, non-sparse regularization is imposed on the representation coefficients. In particular, the proposed RRC model has two important cases with two specific values of $\beta$.

When $\beta=2$, GGD degenerates to the Gaussian distribution, and the RRC model becomes

$$\hat{\alpha} = \arg\min_\alpha \left\{ \left\| W^{1/2}(y - D\alpha) \right\|_2^2 + \lambda \|\alpha\|_2^2 \right\} \tag{18}$$

In this case the RRC model is essentially an $l_2$-regularized robust coding model. It can be easily derived that when $W$ is given, the solution to Eq. (18) is $\hat{\alpha} = (D^T W D + \lambda I)^{-1} D^T W y$.

When $\beta=1$, GGD degenerates to the Laplacian distribution, and the RRC model becomes

$$\hat{\boldsymbol{\alpha}} = \arg\min_{\boldsymbol{\alpha}} \left\{ \left\| W^{1/2} (\boldsymbol{y} - \boldsymbol{D}\boldsymbol{\alpha}) \right\|_2^2 + \lambda \left\| \boldsymbol{\alpha} \right\|_1 \right\} \quad (19)$$

In this case the RRC model is essentially the RSC model in [35], where the sparse coding methods such as $l_1\_ls$ [41] is used to solve Eq. (19) when $W$ is given. In this paper, we solve Eq. (19) via Eq. (17) by the iteratively re-weighting technique [30]. Let $V_{j,j}^{(0)} = v_o^{(0)} = 1$, and then in the $(k+1)^{th}$ iteration $V$ is set as $V_{j,j}^{(k+1)} = v_o(\alpha_j^{(k)}) = \lambda \left| (\alpha_j^{(k)})^2 + \varepsilon^2 \right|^{-1/2}$, and then $\hat{\boldsymbol{\alpha}}^{(k+1)} = \left( \boldsymbol{D}^T W \boldsymbol{D} + V^{(k+1)} \right)^{-1} \boldsymbol{D}^T W \boldsymbol{y}$. Here $\varepsilon$ is a scalar defined in [30].

## 3. Algorithm of RRC

### 3.1. Iteratively reweighted regularized robust coding (IR³C) algorithm

**Table 1:** Algorithm of Iteratively Reweighted Regularized Robust Coding.

| Iteratively Reweighted Regularized Robust Coding (IR³C) |
|---|
| **Input:** Normalized query image $y$ with unit $l_2$-norm; dictionary $D$ (each column of $D$ has unit $l_2$-norm); $\boldsymbol{\alpha}^{(1)}$.<br>**Output:** $\boldsymbol{\alpha}$<br>Start from $t=1$:<br>1. Compute residual $\boldsymbol{e}^{(t)} = \boldsymbol{y} - \boldsymbol{D}\boldsymbol{\alpha}^{(t)}$.<br>2. Estimate weights as $$\omega_{\boldsymbol{\theta}}(e_i^{(t)}) = 1 / 1 + \exp\left( \mu (e_i^{(t)})^2 - \mu\delta \right),$$ where $\mu$ and $\delta$ could be estimated in each iteration (please refer to Section 4.1 for the settings of them).<br>3. Weighted regularized robust coding: $$\boldsymbol{\alpha}^* = \arg\min_{\boldsymbol{\alpha}} \left\{ \frac{1}{2} \left\| (W^{(t)})^{1/2} (\boldsymbol{y} - \boldsymbol{D}\boldsymbol{\alpha}) \right\|_2^2 + \sum_{j=1}^m \rho_o(\alpha_j) \right\} \quad (21)$$ where $W^{(t)}$ is the estimated diagonal weight matrix with $W_{i,i}^{(t)} = \omega_{\boldsymbol{\theta}}(e_i^{(t)})$, $\rho_o(\alpha_j) = \lambda |\alpha_j|^\beta + b_{\alpha_0}$ and $\beta = 2$ or 1.<br>4. Update the sparse coding coefficients:<br>    If $t=1$, $\boldsymbol{\alpha}^{(t)} = \boldsymbol{\alpha}^*$;<br>    If $t>1$, $\boldsymbol{\alpha}^{(t)} = \boldsymbol{\alpha}^{(t-1)} + \upsilon^{(t)} (\boldsymbol{\alpha}^* - \boldsymbol{\alpha}^{(t-1)})$;<br>    where $0 < \upsilon^{(t)} \leq 1$ is a suitable step size that makes $\sum_{i=1}^n \rho_{\boldsymbol{\theta}}(y_i - r_i\boldsymbol{\alpha}^{(t)}) + \sum_{j=1}^m \rho_o(\alpha_j^{(t)}) < \sum_{i=1}^n \rho_{\boldsymbol{\theta}}(y_i - r_i\boldsymbol{\alpha}^{(t-1)}) + \sum_{j=1}^m \rho_o(\alpha_j^{(t-1)})$. $\upsilon^{(t)}$ can be searched from 1 to 0 by the standard line-search process [42].<br>5. Compute the reconstructed test sample: $$\boldsymbol{y}_{rec}^{(t)} = \boldsymbol{D}\boldsymbol{\alpha}^{(t)},$$ and let $t=t+1$.<br>6. Go back to step 1 until the condition of convergence (refer to Section 3.2) is met, or the maximal number of iterations is reached. |

As discussed in Section 2, the minimization of RRC is an iterative process, and the weights $W$ and $V$ are updated alternatively in order for the desired coding vector $\boldsymbol{\alpha}$. Although we can only have a locally optimal solution to the RRC model, fortunately in FR we are able to have a very reasonable initialization to achieve good performance. In this section we propose an iteratively reweighted regularized robust coding (IR$^3$C) algorithm to minimize the RRC model.

When a query face image $y$ comes, in order to initialize $W$, we should firstly initialize the coding residual $e$ of $y$. We initialize $e$ as $e=y-D\alpha^{(1)}$, where $\alpha^{(1)}$ is an initial coding vector. Because we do not know which class the query face image $y$ belongs to, a reasonable $\alpha^{(1)}$ can be set as

$$\alpha^{(1)} = \left[\tfrac{1}{m}; \tfrac{1}{m}; \ldots; \tfrac{1}{m}\right] \tag{20}$$

That is, $D\alpha^{(1)}$ is the mean image of all training samples. With the initialized coding vector $\alpha^{(1)}$, the proposed IR$^3$C algorithm is summarized in Table 1.

When IR$^3$C converges, we use the same classification strategy as in SRC [18] to classify the face image $y$:

$$\text{identity}(y) = \arg\min_c \{\ell_c\} \tag{22}$$

where $\ell_c = \left\| W_{final}^{1/2} (y - D_c \hat{\alpha}_c) \right\|_2$, $D_c$ is the sub-dictionary associated with class $c$, $\hat{\alpha}_c$ is the final sub-coding vector associated with class $c$, and $W_{final}$ is the final weight matrix.

## 3.2. The convergence of IR$^3$C

Eq. (21) is a local approximation of the RRC in Eq. (8), and in each iteration the objective function of Eq. (8) decreases by the IR$^3$C algorithm, i.e., in steps 3 and 4, the solved $\alpha^{(t)}$ will make $\sum_{i=1}^{n} \rho_\theta \left( y_i - r_i \alpha^{(t)} \right) + \sum_{j=1}^{m} \rho_o \left( \alpha_j^{(t)} \right) < \sum_{i=1}^{n} \rho_\theta \left( y_i - r_i \alpha^{(t-1)} \right) + \sum_{j=1}^{m} \rho_o \left( \alpha_j^{(t-1)} \right)$. Since the cost function of Eq. (8) is lower bounded ($\geq 0$), the iterative minimization procedure in IR$^3$C will converge. Specifically, we stop the iteration if the following holds:

$$\left\| W^{(t+1)} - W^{(t)} \right\|_2 / \left\| W^{(t)} \right\|_2 < \delta_W \tag{23}$$

where $\delta_W$ is a small positive scalar.

## 3.3. Complexity analysis

Generally speaking, the complexity of IR$^3$C and SRC [18] mainly lies in the coding process, i.e., Eq. (18) or (19) for IR$^3$C and Eq. (1) or Eq. (2) for SRC. It is known that the $l_1$-minimization, such as Eq. (1) for SRC, has a computational complexity of $O(n^2 m^{1.5})$ [51], where $n$ is the dimensionality of face feature, and $m$ is the number of dictionary atoms. It is also reported that the commonly used $l_1$-minmization solvers, e.g., $l_1$\_magic [43] and $l_1\_ls$ [41], have an empirical complexity of $O(n^2 m^{1.3})$ [41].

For IR$^3$C with $\beta$=2, the coding (i.e., Eq. (18)) is an $l_2$-regularized least square problem. The solution $\hat{\boldsymbol{\alpha}} = \left(\boldsymbol{D}^T \boldsymbol{W} \boldsymbol{D} + \lambda \boldsymbol{I}\right)^{-1} \boldsymbol{D}^T \boldsymbol{W} \boldsymbol{y}$ could be got by solving $\left(\boldsymbol{D}^T \boldsymbol{W} \boldsymbol{D} + \lambda \boldsymbol{I}\right) \hat{\boldsymbol{\alpha}} = \boldsymbol{D}^T \boldsymbol{W} \boldsymbol{y}$ efficiently via conjugate gradient method [50], whose time complexity is about $O(k_1 nm)$ (here $k_1$ is the iteration number in conjugate gradient method). Suppose that $t$ iterations are used in IR$^3$C to update $\boldsymbol{W}$, the overall complexity of IR$^3$C with $\beta$=2 is about $O(tk_1 nm)$. Usually $t$ is less than 15. It is easy to see that IR$^3$C with $\beta$=2 has much lower complexity than SRC.

For IR$^3$C with $\beta$=1, the coding in Eq. (19) is an $l_1$-norm sparse coding problem, which could also be solved via conjugate gradient method. The complexity of IR$^3$C with $\beta$=1 will be about $O(tk_1 k_2 nm)$, where $k_2$ is the number of iteration to update $\boldsymbol{V}$. By experience, $k_1$ is less than 30 and $k_2$ is less 20, and then $k_2 k_1$ is basically in the similar order to $n$. Thus the complexity of IR$^3$C with $\beta$=1 is about $O(tn^2 m)$. Compared with SRC in case of FR without occlusion, although IR$^3$C needs several iterations (usually $t$=2) to update $\boldsymbol{W}$, its time consumption is still lower than or comparable to SRC. In FR with occlusion or corruption, for IR$^3$C usually $t$=15. In this case, however, SRC's complexity is $O(n^2(m+n)^{1.3})$ because it needs to use an identity matrix to code the occluded or corrupted pixels, as shown in Eq. (2). It is easy to conclude that IR$^3$C with $\beta$=1 has much lower complexity than SRC for FR with occlusion.

Although many faster $l_1$-norm minimization methods than $l_1\_magic$ [43] and $l_1\_ls$ [41] have been proposed recently, as reviewed in [60], by adopting them in SRC the running time is still larger than or comparable to the proposed IR$^3$C, as demonstrated in Section 4.5. In addition, in the iteration of IR$^3$C we can delete the element $y_i$ that has very small weight because this implies that $y_i$ is an outlier. Thus the complexity of IR$^3$C can be further reduced. For example, in FR with real disguise on the AR database, about 30% pixels could be deleted.

## 4. Experimental Results

We perform experiments on benchmark face databases to demonstrate the performance of RRC. In Section 4.1, we give the parameter setting of RRC; in Section 4.2, we test RRC for FR without occlusion; in Section 4.3, we demonstrate the robustness of RRC to FR with random pixel corruption, random block occlusion and real disguise; in Section 4.4, the experiments on rejecting invalid testing images are performed. In Section 4.5, the running time is presented. Finally, some discussions of parameter selection are given in Section 4.6.

All the face images are cropped and aligned by using the locations of eyes. We normalize the query image (or feature) and training image (or feature) to have unit $l_2$-norm energy. For AR [44] and Extended Yale B [13, 45] databases, the eye locations are provided by the databases. For Multi-PIE [46] database, we manually locate the eyes for the experiments in Section 4.2, and automatically detect the facial region by the face detector [47] for the experiments in Sections 4.4. In all experiments, the training samples are used as the dictionary $D$ in coding. We denote by RRC_L$_1$ our RRC model with $l_1$-norm coefficient constraint (i.e., $\beta$=1 in Eq. (19)), and by RRC_L$_2$ our RRC model with $l_2$-norm coefficient constraint (i.e., $\beta$=2 in Eq. (18)). Both RRC_L$_1$ and RRC_L$_2$ are implemented by the IR$^3$C algorithm described in Section 3.1.

**4.1. Parameter setting**

In the weight function Eq. (14), there are two parameters, $\delta$ and $\mu$, which need to be calculated in Step 2 of the IR$^3$C algorithm. $\delta$ is the parameter of demarcation point. When the square of residual is larger than $\delta$, the weight will be less than 0.5. To make the model robust to outliers, we compute $\delta$ as follows. Let $l=\lfloor \tau n \rfloor$, where scalar $\tau \in (0,1)$, and $\lfloor \tau n \rfloor$ outputs the largest integer smaller than $\tau n$. We set $\delta$ as

$$\delta = \psi_1(\boldsymbol{e})_l \tag{24}$$

where for a vector $\boldsymbol{e} \in \Re^n$, $\psi_1(\boldsymbol{e})_k$ is the $k^{\text{th}}$ largest element of the set $\{e_j^2, j=1,\ldots,n\}$.

Parameter $\mu$ controls the decreasing rate of weight $W_{i,i}$. Here we simply let $\mu=\varsigma/\delta$, where $\varsigma=8$ is set as a constant. In the experiments, $\tau$ is fixed as 0.8 for FR without occlusion, and 0.6 for FR with occlusion. In addition, the regularization parameter $\lambda$ in Eq. (18) or Eq. (19) is set as 0.001 by default.

For RRC_L$_1$, there is a parameter $\varepsilon$ in updating the weight matrix $V$: $V_{j,j}^{(k+1)} = v_o\left(\alpha_j^{(k)}\right) = b_o \left| (\alpha_j^{(k)})^2 + \varepsilon^2 \right|^{-1/2}$. According to [30], we choose $\varepsilon$ as

$$\varepsilon^{(k+1)} = \min\left(\varepsilon^{(k)}, \psi_2\left(\boldsymbol{\alpha}^{(k)}\right)_L / m\right) \tag{25}$$

where for a vector $\boldsymbol{\alpha} \in \Re^m$, $\psi_2(\boldsymbol{\alpha})_i$ is the $i^{\text{th}}$ largest element of the set $\{|\alpha_j|, j=1,\cdots,m\}$. We set $L=\lfloor 0.01m \rfloor$. The above design of $\varepsilon$ could not only make the numerical computing of weight $V$ stable, but also ensure the iteratively reweighted least square achieve a sparse solution ($\varepsilon^{(k+1)}$ decreases to zero as $k$ increases).

## 4.2. Face recognition without occlusion

We first validate the performance of RRC in FR with variations such as illumination and expression changes but without occlusion. We compare RRC with SRC [18], locality-constrained linear coding (LLC) [31], linear regression for classification (LRC) [15] and the benchmark methods such as nearest neighbor (NN), nearest feature line (NFL) [10] and linear support vector machine (SVM). In the experiments, PCA is used to reduce the dimensionality of original face images, and the Eigenface features are used for all the competing methods. Denote by $P$ the PCA projection matrix, the step 3 of IR$^3$C becomes:

$$\alpha^* = \arg\min_{\alpha} \left\{ \frac{1}{2} \left\| P(W^{(t)})^{1/2} (y - D\alpha) \right\|_2^2 + \sum_{j=1}^{m} \rho_o(\alpha_j) \right\} \tag{26}$$

**Table 2:** Face recognition rates on the Extended Yale B database.

| Dimension | 30 | 84 | 150 | 300 |
|---|---|---|---|---|
| NN | 66.3% | 85.8% | 90.0% | 91.6% |
| SVM | **92.4%** | 94.9% | 96.4% | 97.0% |
| LRC [15] | 63.6% | 94.5% | 95.1% | 96.0% |
| NFL [10] | 89.6% | 94.1% | 94.5% | 94.9% |
| SRC [18] | 90.9% | 95.5% | 96.8% | 98.3% |
| LLC [31] | 92.1% | 96.4% | 97.0% | 97.6% |
| RRC_L$_2$ | 71.6% | 94.4% | 97.6% | 98.9% |
| RRC_L$_1$ | 91.3% | **98.0%** | **98.8%** | **99.8%** |

**Table 3:** Face recognition rates on the AR database.

| Dimension | 30 | 54 | 120 | 300 |
|---|---|---|---|---|
| NN | 62.5% | 68.0% | 70.1% | 71.3% |
| SVM | 66.1% | 69.4% | 74.5% | 75.4% |
| LRC [15] | 66.1% | 70.1% | 75.4% | 76.0% |
| NFL [10] | 64.5% | 69.2% | 72.7% | 73.4% |
| SRC [18] | **73.5%** | 83.3% | 90.1% | 93.3% |
| LLC [31] | 70.5% | 80.7% | 87.4% | 89.0% |
| RRC_L$_2$ | 61.5% | 84.3% | 94.3% | 95.3% |
| RRC_L$_1$ | 70.8% | **87.6%** | **94.7%** | **96.3%** |

*4.2.1) Extended Yale B Database:* The Extended Yale B [13, 45] database contains about 2,414 frontal face images of 38 individuals. We used the cropped and normalized face images of size 54×48, which were taken under varying illumination conditions. We randomly split the database into two halves. One half, which contains 32 images for each person, was used as the dictionary, and the other half was used for testing. Table 2 shows the recognition rates versus feature dimension by NN, NFL, SVM, SRC, LRC, LLC and RRC methods. RRC_L$_1$ achieves better results than the other methods in all dimensions except that they are slightly worse than SVM when the dimension is 30. RRC_L$_2$ is better than SRC, LRC, LLC, SVM, NFL and NN when the dimension is 150 or higher. The best recognition rates of SVM, SRC, LRC, LLC, RRC_$L_2$ and RRC_$L_1$ are 97.0%, 98.3%, 96.0%, 97.6%, 98.9% and 99.8% respectively.

*4.2.2) AR Database:* As in [18], a subset (with only illumination and expression changes) that contains 50 male and 50 female subjects was chosen from the AR database [44] in this experiment. For each subject, the seven images from Session 1 were used for training, with other seven images from Session 2 for testing. The images were cropped to 60×43. The FR rates by the competing methods are listed in Table 3. We can see that apart from the case when the dimension is 30, RRC_$L_1$ achieves the highest rates among all methods, while RRC_$L_2$ is the second best. The reason that RRC works not very well with very low-dimensional feature is that the coding vector solved by Eq. (26) is not accurate enough to estimate $W$ when the feature dimension is too low. Nevertheless, when the dimension is too low, all the methods cannot achieve good recognition rate. We can see that all methods achieve their maximal recognition rates at the dimension of 300, with 93.3% for SRC, 89.0% for LLC, 95.3% for RRC_$L_2$ and 96.3% for RRC_$L_1$.

From Table 2 and Table 3, one can see that when the dimension of feature is not too low, RRC_$L_2$ could achieve similar performance to that of RRC_$L_1$, which implies that the $l_1$-sparsity constraint on the coding vector is not so important. This is because when the feature dimension is not too low, the dictionary (i.e., the feature set of the training samples) may not be over-complete enough, and hence using Laplacian to model the coding vector is not much better than using Gaussian. As a result, RRC_$L_2$ and RRC_$L_1$ will have similar recognition rates, but the former will have much less complexity.

*4.2.3) Multi PIE database:* The CMU Multi-PIE database [46] contains images of 337 subjects captured in four sessions with simultaneous variations in pose, expression, and illumination. Among these 337 subjects, all the 249 subjects in Session 1 were used for training. To make the FR more challenging, four subsets with both illumination and expression variations in Sessions 1, 2 and 3, were used for testing. For the training set, as in [28], we used the 7 frontal images with extreme illuminations {0, 1, 7, 13, 14, 16, and 18} and neutral expression (refer to Fig. 2(a) for examples). For the testing set, 4 typical frontal images with illuminations {0, 2, 7, 13} and different expressions (smile in Sessions 1 and 3, squint and surprise in Session 2) were used (refer to Fig. 2(b) for examples with surprise in Session 2, Fig. 2(c) for examples with smile in Session 1, and Fig. 2(d) for examples with smile in Session 3). Here we used the Eigenface with dimensionality 300 as the face feature for sparse coding. Table 4 lists the recognition rates in four testing sets by the competing methods.

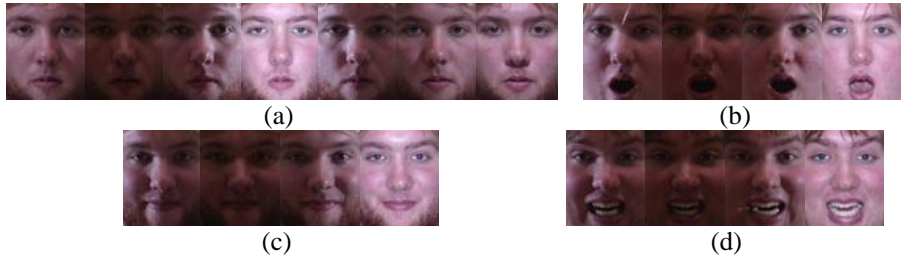

(a)                                (b)

(c)                                (d)

**Figure 2:** A subject in Multi-PIE database. (a) Training samples with only illumination variations. (b) Testing samples with surprise expression and illumination variations. (c) and (d) show the testing samples with smile expression and illumination variations in Session 1 and Session 3, respectively.

**Table 4:** Face recognition rates on Multi-PIE database. ('Smi-S1': set with smile in Session 1; 'Smi-S3': set with smile in Session 3; 'Sur-S2': set with surprise in Session 2; 'Squ-S2': set with squint in Session 2).

|  | Smi-S1 | Smi-S3 | Sur-S2 | Squ-S2 |
| --- | --- | --- | --- | --- |
| NN | 88.7% | 47.3% | 40.1% | 49.6% |
| SVM | 88.9% | 46.3% | 25.6% | 47.7% |
| LRC [15] | 89.6% | 48.8% | 39.6% | 51.2% |
| NFL [10] | 90.3% | 50.0% | 39.8% | 52.9% |
| SRC [18] | 93.7% | 60.3% | 51.4% | 58.1% |
| LLC [31] | 95.6% | 62.5% | 52.3% | 64.0% |
| RRC_$L_2$ | 96.1% | 70.2% | 59.2% | 58.1% |
| RRC_$L_1$ | **97.8%** | **76.0%** | **68.8%** | **65.8%** |

From Table 4, we can see that RRC_$L_1$ achieves the best performance in all tests, and RRC_$L_2$ performs the second best. Compared to the third best method, LLC, 6% and 2.3% average improvements are achieved by RRC_$L_1$ and RRC_$L_2$, respectively. In addition, all the methods achieve their best results when Smi-S1 is used for testing because the training set is also from Session 1. From testing set Smi-S1 to Smi-S3, the variations increase because of the longer data acquisition time interval and the difference of smile (refer to Fig. 2(c) and Fig. 2(d)). The recognition rates of RRC_$L_1$ and RRC_$L_2$ drop by 21.8% and 25.9%, respectively, while those of NN, NFL, LRC, SVM, LLC and SRC drop by 41.4%, 40.3%, 40.8%, 42.6%, 33.1% and 33.4%, respectively. This validates that the RRC methods are much more robust to face variations than the other methods. Meanwhile, we could also see that FR with surprise and squint expression changes are much more difficult than FR with the smile expression change. In this experiment, the gap between RRC_$L_2$ and RRC_$L_1$ is relatively big. The reason is that the dictionary (size: 300×1743) used in this experiment is much over-complete, and thus the $l_1$-norm is much more powerful than the $l_2$-norm to regularize the representation of samples with big variations (e.g., expression changes).

### 4.3. Face recognition with occlusion

One of the most interesting features of sparse coding based FR in [18] is its robustness to face occlusion. In this subsection, we test the robustness of RRC to different kinds of occlusions, such as random pixel corruption, random block occlusion and real disguise. In the experiments of random corruption and random block occlusion, we compare RRC methods with SRC [18], LRC [15], Gabor-SRC [20] (only suitable for block occlusion) and correntropy-based sparse representation (CESR) [33], and NN is used as the baseline method. In the experiment of real disguise, we compare RRC with SRC, Gabor-SRC (GSRC) [20], CESR and other state-of-the-art methods.

*4.3.1) FR with pixel corruption*: To be identical to the experimental settings in [18], we used Subsets 1 and 2 (717 images, normal-to-moderate lighting conditions) of the Extended Yale B database for training, and used Subset 3 (453 images, more extreme lighting conditions) for testing. The images were resized to 96×84 pixels. For each testing image, we replaced a certain percentage of its pixels by uniformly distributed random values within [0, 255]. The corrupted pixels were randomly chosen for each test image and the locations are unknown to the algorithm.

Fig. 3 shows a representative example of RRC_$L_1$ and RRC_$L_2$ with 70% random corruption. Fig. 3(a) is the original sample, and Fig. 3(b) shows the testing image with random corruption. It can be seen that the corrupted face images are difficult to recognize, even for humans. The estimated weight maps of RRC_$L_1$ and RRC_$L_2$ are shown in the top and bottom rows of Fig. 3(c) respectively, from which we can see not only the corrupted pixels but also the pixels in the shadow region have low weights. Fig. 3(d) shows the coding coefficients of RRC_$L_1$ (top row) and RRC_$L_2$ (bottom row), while Fig. 3(e) shows the reconstructed images of RRC_$L_1$ (top row) and RRC_$L_2$ (bottom row). It can be seen that for RRC_$L_1$ only the dictionary atoms with the same label as the testing sample have big coefficients and the reconstructed image is faithful to the original image (Fig. 3(a)) but with better visual quality (the shadow which brings difficulties to recognition is removed). For RRC_$L_2$, although the coefficients are not sparse, the visual quality of the reconstructed image is also good and the classification performance is similar to RRC_$L_1$, which are shown in Table 5.

Table 5 shows the results of SRC, CESR, LRC, NN, RRC_$L_2$ and RRC_$L_1$ under different percentage of corrupted pixels. Since all competing methods could achieve no bad performance from 0% to 50% corruptions, we only list the average recognition rate for 0%~50% corruptions. One can see that when the percentage of corrupted pixels is between 0% and 50%, RRC_$L_1$, RRC_$L_2$, and SRC could correctly classify all the testing images. Surprisingly, CESR does not correctly recognize all the testing images in that case. However, when the percentage of corrupted pixels is more than 70%, the advantage of RRC_L$_1$, RRC_L$_2$, and CESR over SRC is clear. Especially, RRC_L$_1$ achieves the best performance in all cases, with 100% (99.6% and 67.1%) in 70% (80% and 90%) corruption, while SRC only has a recognition rate of 90.7% (37.5% and 7.1%). LRC and NN are sensitive to the outliers, with much lower recognition rates than others. All RRC methods achieve better performance than CESR in all cases, which validates that the RRC model could suppress the effect of outliers better. Meanwhile, we see that RRC_L$_2$ has very similar performance to RRC_L$_1$, which shows that when the feature dimension (8064 here) is high, $l_2$-norm constraint on coding coefficient is as powerful as $l_1$-norm constraint, but with much less time complexity.

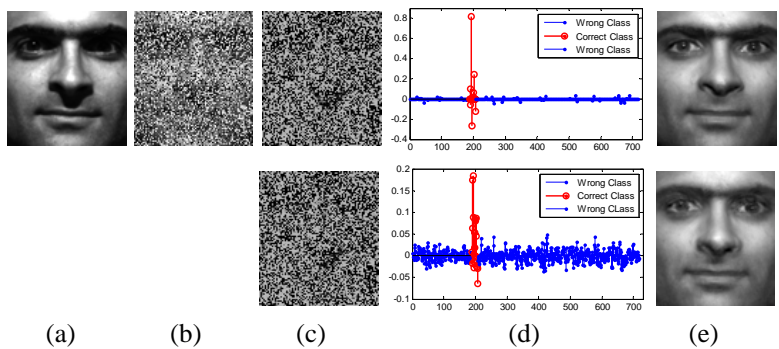

(a) (b) (c) (d) (e)

**Figure 3**: Recognition under random corruption. (a) Original image $y_0$ from Extended Yale B database. (b) Test image $y$ with random corruption. (c) Estimated weight map of RRC_L$_1$ (top row) and RRC_L$_2$ (bottom). (d) Estimated representation coefficients $\alpha$ of RRC_L$_1$ and RRC_L$_2$. (e) Reconstructed images $y_{rec}$ of RRC_L$_1$ and RRC_L$_2$.

**Table 5:** The recognition rates of RRC, LRC, NN, SRC and CESR versus different percentage of corruption.

| Corruption(%) | 0~50 (average) | 60 | 70 | 80 | 90 |
|---|---|---|---|---|---|
| NN | 89.3% | 46.8% | 25.4% | 11.0% | 4.6% |
| SRC [18] | **100%** | 99.3% | 90.7% | 37.5% | 7.1% |
| LRC [15] | 95.8% | 50.3% | 26.4% | 9.9% | 6.2% |
| CESR [33] | 97.4% | 96.2% | 97.8% | 93.8% | 41.5% |
| RRC_L$_2$ | **100%** | **100%** | 99.8% | 97.8% | 43.3% |
| RRC_L$_1$ | **100%** | **100%** | **100%** | 99.6% | 67.1% |

*4.3.2) FR with block occlusion*: In this section we test the robustness of RRC to block occlusion. We also used the same experimental settings as in [18], i.e., Subsets 1 and 2 of Extended Yale B for training, Subset 3 for testing, and replacing a randomly located square block of a test image with an unrelated image, as illustrated in Fig. 4(b). The face images were resized to 96×84.

Fig. 4 shows an example of occluded face recognition (30% occlusion) by using RRC_L$_1$ and RRC_L$_2$. Fig. 4 (a) and (b) are the original sample from Extended Yale B database and the occluded testing sample. Fig. 4 (c) shows the estimated weight maps of RRC_$L_1$ (top row) and RRC_$L_2$ (bottom row), from which we could see that both of them assign big weights (e.g., 1) to the un-occluded pixels, and assign low weight (e.g., 0) to the occluded pixels. The estimated representation coefficients of RRC_$L_1$ and RRC_$L_2$ are shown in the top row and bottom row of Fig. 4 (d) respectively. It can be seen that RRC_$L_1$ could achieve very sparse coefficients with significant values on the atoms of correct class; the coefficients by RRC_$L_2$ also have significant values on the atoms of correct class but they are not sparse. From Fig. 4 (e), we see that both RRC_$L_1$ and RRC_$L_2$ have very good image reconstruction quality, effectively removing the block occlusion and the shadow.

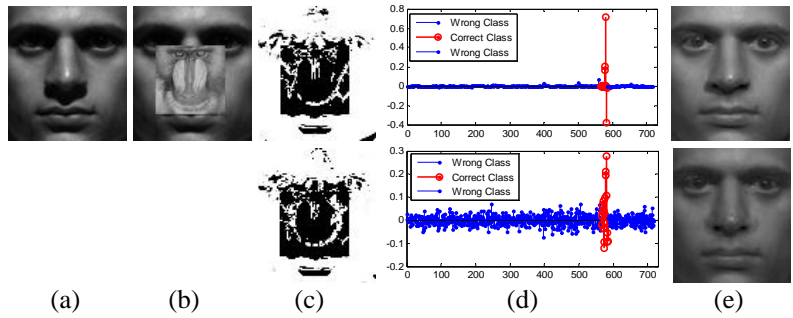

(a) (b) (c) (d) (e)

**Figure 4:** Recognition under 30% block occlusion. (a) Original image $y_0$ from Extended Yale B. (b) Test image $y$ with random corruption. (c) Estimated weight maps of RRC_$L_1$ (top row) and RRC_$L_2$ (bottom row). (d) Estimated representation coefficients $\alpha$ of RRC_$L_1$ and RRC_$L_2$. (e) Reconstructed images $y_{rec}$ of RRC_$L_1$ and RRC_$L_2$.

**Table 6:** The recognition rates of RRC, LRC, NN, GSRC, SRC and CESR under different levels of block occlusion.

| Occlusion (%) | 0 | 10 | 20 | 30 | 40 | 50 |
|---|---|---|---|---|---|---|
| NN | 94.0% | 92.9% | 85.4% | 73.7% | 62.9% | 45.7% |
| SRC [18] | **100%** | **100%** | 99.8% | 98.5% | 90.3% | 65.3% |
| LRC [15] | **100%** | **100%** | 95.8% | 81.0% | 63.8% | 44.8% |
| GSRC[20] | **100%** | **100%** | **100%** | **99.8%** | 96.5% | 87.4% |
| CESR[33] | 94.7% | 94.7% | 92.7% | 89.8% | 83.9% | 75.5% |
| RRC_$L_2$ | **100%** | **100%** | **100%** | **99.8%** | **97.6%** | **87.8%** |
| RRC_$L_1$ | **100%** | **100%** | **100%** | **99.8%** | 96.7% | 87.4% |

Table 6 lists the detailed recognition rates of RRC_$L_1$, RRC_$L_2$, SRC, LRC, NN, GSRC and CESR under the occlusion percentage from 0% to 50%. From Table 6, we see that RRC_$L_2$ has the best accuracy, and RRC methods achieve much higher recognition rates than SRC when the occlusion percentage is larger than 30% (e.g., more than 22% (6%) improvement at 50% (40%) occlusion). Compared to GSRC, RRC still gets better results without using the enhanced Gabor features. CESR gets worse results than SRC in this experiment. This may be because FR with block occlusion is more difficult than that of pixel corruption, but it shows that CESR could not accurately identify the outlier points in such block occlusion (i.e., outlier points have similar intensity as the face pixels). Encouragingly, RRC_L$_2$ also has competing recognition rates to RRC_L$_1$ (even better than them at 40% and 50% occlusion), which validates that the low-complexity $l_2$-norm regularization could be as powerful as the $l_1$-norm regularization for such kind of block occlusions.

*4.3.4) FR with real face disguise:* A subset from the AR database is used in this experiment. This subset consists of 2,599 images from 100 subjects (26 samples per class except for a corrupted image w-027-14.bmp), 50 males and 50 females. We perform two tests: one follows the experimental settings in [18], while the other one is more challenging. The images were resized to 83×60 in the first test and 42×30 in the second test.

In the first test, 799 images (about 8 samples per subject) of non-occluded frontal views with various facial expressions in Sessions 1 and 2 were used for training, while two separate subsets (with sunglasses and scarf) of 200 images (1 sample per subject per Session, with neutral expression) for testing. Fig. 5 illustrates the classification process of RRC_L$_1$ by using an example. Fig. 5(a) shows a test image with sunglasses; Figs. 5(b) and 5(c) show the initialized and final weight maps, respectively; Fig. 5(d) shows one template image of the identified subject. The convergence of the IR$^3$C algorithm to solve the RRC model is shown in Fig. 5(e), and Fig. 5(f) shows the reconstruction error of each class, with the correct class having the lowest value. The FR results by the competing methods are listed in Table 7. We see that the RRC methods achieve much higher recognition rates than SRC, GSRC and CESR, while RRC_L$_1$ and RRC_L$_2$ achieve similar results. CESR has similar performance to RRC methods in FR with sunglass, but has much worse recognition rate in dealing with scarf. Similar to the case of FR with block occlusion, CESR is not robust enough for more challenging case (i.e., scarf covers about 40% face region). The proposed RRC methods also significantly outperform other state-of-the-art methods, including [48] with 84% on sunglasses and 93% on scarf, and [26] with 93% on sunglasses and 95.5% on scarf.

In the second test, we conduct FR with more complex disguise (disguise with variations of illumination and longer data acquisition interval). 400 images (4 neutral images with different illuminations per subject) of non-occluded frontal views in Session 1 were used for training, while the disguised images (3 images with various illuminations and sunglasses or scarves per subject per Session) in Sessions 1 and 2 for testing. Table 8 lists the results by competing methods. Clearly, the RRC methods achieve much better results than SRC, GSRC and CESR. Interestingly, CESR works well in the case of Sunglasses disguise but poor in the case of Scarves disguise, while GSRC the reverse. In addition, the average improvements of RRC_L$_1$ over SRC,

GSRC and CESR are respectively 21.4%, 28% and 7% on sunglasses, and respectively 62.3%, 9.3% and 55.5% on scarf. In this experiment, RRC_$L_1$ is slightly better than RRC_$L_2$ on sunglasses, with RRC_$L_2$ slightly better than RRC_$L_1$ on scarf.

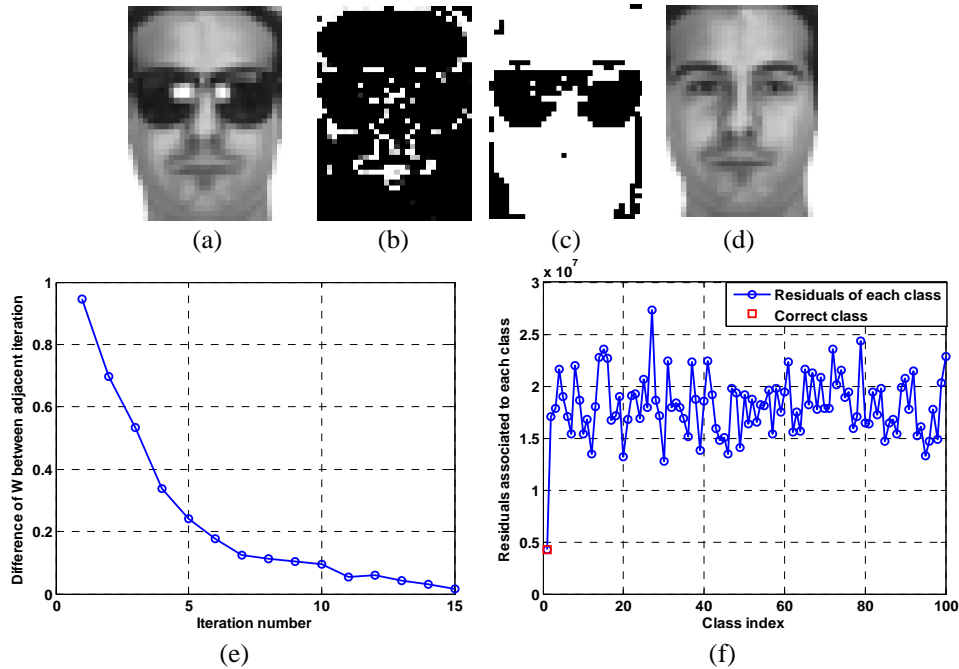

**Figure 5:** An example of face recognition with disguise using RRC_$L_1$. (a) A test image with sunglasses. (b) The initialized weight map. (c) The weight map when IR$^3$C converges. (d) A template image of the identified subject. (e) The convergence curve of IR$^3$C. (f) The residuals of each class by RRC_$L_1$.

**Table 7:** Recognition rates by competing methods on the AR database with disguise occlusion.

| Algorithms | Sunglasses | Scarves |
| --- | --- | --- |
| SRC [18] | 87.0% | 59.5% |
| GSRC [20] | 93% | 79% |
| CESR[33] | 99% | 42.0% |
| RRC_$L_2$ | 99.5% | 96.5% |
| RRC_$L_1$ | **100%** | **97.5%** |

**Table 8:** Recognition rates by competing methods on the AR database with complex disguise occlusion.

| Algorithms | Session 1 | | Session 2 | |
| --- | --- | --- | --- | --- |
|  | Sunglasses | Scarves | Sunglasses | Scarves |
| SRC [18] | 89.3% | 32.3% | 57.3% | 12.7% |
| GSRC [20] | 87.3% | 85% | 45% | 66% |
| CESR[33] | 95.3% | 38% | 79% | 20.7% |
| RRC_$L_2$ | **99.0%** | **94.7%** | 84.0% | **77.3%** |
| RRC_$L_1$ | **99.0%** | 93.3% | **89.3%** | 76.3% |

## 4.4. Face validation

In practical FR systems, it is important to reject invalid face images which have no template in the database. It should be noted that "*rejecting invalid images not in the entire database is much more difficult than deciding if two face images are the same subject*" [28]. In this section we check whether the proposed RRC methods could have good face validation performance. Similar to [18, 28], all the competing methods use the *Sparsity Concentration Index* (SCI) proposed in [18] to do face validation with the coding coefficient. Like [28], we used the large-scale Multi-PIE face database to perform face validation experiments. All the 249 subjects in Session 1 were used as the training set, with the same subjects in Session 2 as customer images. The remaining 88 subjects (37 subjects with ID between 251 and 292 from Session 2 and 51 subjects with ID between 293 and 346 from Session 3) different from the training set were used as the imposter images. For the training set, as in [28] we used the 7 frontal images with extreme illuminations {0, 1, 7, 13, 14, 16, and 18} and neutral expression (refer to Fig. 2(a) for examples). For the testing set, 10 typical frontal images of illuminations {0, 2, 4, 6, 8, 10, 12, 14, 16, 18} taken with neutral expressions were used. In this experiment, the testing face images were automatically detected by using Viola and Jones' face detector [47] and then automatically aligned to the size of 60×48 without manual intervention (a testing image is automatically aligned to the training data of each subject by the method in [28]).

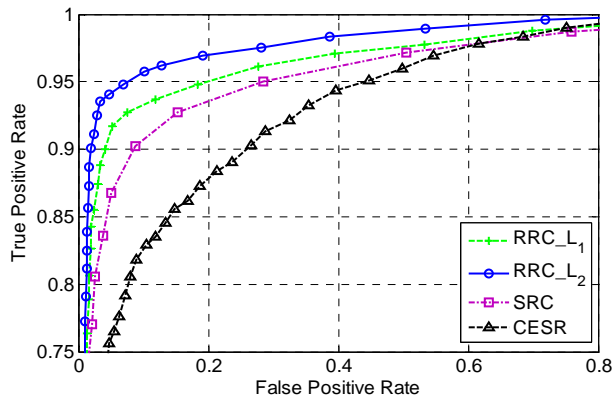

**Figure 6:** Subject validation on the large-scale Multi PIE.

Fig. 6 plots the ROC (receiver operating characteristic) curves of the competing methods: SRC, RRC_$L_1$, RRC_$L_2$ and CESR. It can be seen that CESR works the worst while RRC_$L_2$ works the best. For instance, when the false positive rate is 0.1, the true positive rate is 82.6% for CESR, 90.7% for SRC, 93.3% for RRC_$L_1$ and 95.8% for RRC_$L_2$. It is a little surprising that RRC_$L_2$ with $l_2$-norm coefficient constraint achieves better face validation results than the $l_1$-norm coefficient constrained methods, e.g., SRC, RRC_L$_1$, and much better than CESR. The reason may be that the $l_1$-norm constraint, especially the nonnegative sparse constraint (for CESR), which strongly forces the coding coefficients to be sparse, will force one specific class to represent the input invalid testing sample, and hence incorrectly recognize this testing sample. Comparatively, $l_2$-norm constraint does not force the coding coefficients to be sparse, which allows the representation coefficients of invalid testing samples to be evenly distributed across different classes. Therefore the incorrect recognition can be avoided. In addition, RRC_L$_1$ are better than SRC and CESR, validating that the signal fidelity term of RRC_L$_1$ is more robust.

### 4.5. Running time comparison

Apart from recognition rate, computational expense is also an important issue for practical FR systems. In this section, the running time of the competing methods, including SRC, GSRC, CESR, RRC_L$_2$ and RRC_L$_1$, is evaluated using two FR experiments (without occlusion and with real disguise). The programming environment is Matlab version 7.0a. The desktop used is of 3.16 GHz CPU and with 3.25G RAM. All the methods are implemented using the codes provided by the authors. For SRC, we adopt $l_1\_ls$ [41], and two fast $l_1$-minimization solvers, ALM [60] and Homotopy [61], to implement the sparse coding step.

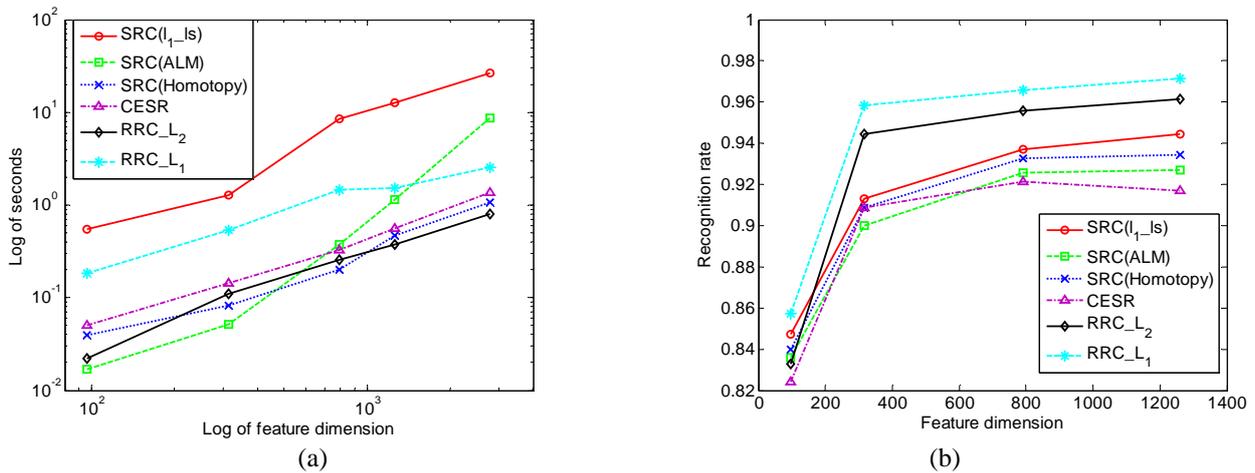

**Figure 7:** Running time and recognition rates by the competing methods under different feature dimension in FR without occlusion.

The first experiment is FR without occlusion on the AR database, whose experimental setting is the same as that in Section 4.2 but with various down-sampled face features (i.e., 12×8, 21×15, 33×24, 42×30 and 62×45). Fig. 7 compares the running time (Fig. 7 (a)) and recognition rates (Fig. 7 (b)) of the competing methods under various feature dimensions. From Fig. 7 (a), it can be seen that RRC_$L_2$, CESR and SRC (Homotopy) have obvious faster speed than other methods. RRC_$L_1$ is also much more efficient than SRC ($l_1\_ls$), the slowest one.

With the feature of 792 (33×24) dimensions, RRC_$L_2$, CESR, RRC_$L_1$, SRC ($l_1\_ls$), SRC (ALM) and SRC (Homotopy) take 0.257, 0.330, 1.450, 8.551, 0.377 and 0.199 seconds, respectively. RRC_$L_1$ achieves the best recognition rates followed by RRC_$L_2$, as shown in Fig. 7(b). Although CESR is also fast, its recognition rates are lower than other methods. It can be concluded that compared to SRC and CESR, RRC_$L_2$ has good recognition rate with much less or comparable computation expense, while RRC_$L_1$ has much higher recognition rate.

The second experiment is FR with real face disguise. The experimental settings are described in Section 4.3. The dictionary has 800 training samples with size 83×60 in Test 1, and 400 training samples with size 42×30 in Test 2. The recognition rates have been reported in Table 7 (for Test 1) and Table 8 (for Test 2). Table 9 lists the average computational expense and recognition rates of different methods on Test1 and Test2. Clearly, RRC_$L_2$ has the least computation time, followed by CESR and RRC_$L_1$. SRC has rather high computation burden even with fast solvers such as ALM and Homotopy, which is because an additional identity matrix is utilized to code occlusion. For the recognition rate, SRC's performance is the worst, and CESR also has rather bad recognition rate in FR with scarf in each test. GSRC solved by $l_1\_ls$ has lower time cost than SRC ($l_1\_ls$) but still very slow. Considering both the recognition rate and running time, RRC_$L_1$ and RRC_$L_2$ are the best ones. RRC_$L_1$ gets the highest recognition rates in almost all cases, at the same time with faster speed than SRC and GSRC. RRC_$L_2$ is the fastest one in all case, at the same time with the second best performance (e.g., in the Test 2 of FR with scarf, 63.5%, 10.5% and 56.6% higher than SRC($l_1\_ls$), GSRC, and CESR in average).

**Table 9:** The average running time (seconds) of competing methods in FR with real face disguise. The values in parenthesis are the average recognition rate.

| Method | Test 1-sunglass | Test 1-scarf | Test 2-sunglass | Test 2- scarf |
|---|---|---|---|---|
| CESR[33] | 2.50 (99.0%) | 3.61 (42.0%) | 0.45 (87.2%) | 0.47 (29.4%) |
| SRC($l_1\_ls$) | 662.15 (87.0%) | 727.14 (59.5%) | 38.23 (73.3%) | 47.73 (22.5%) |
| SRC(ALM) | 35.99 (84.5%) | 36.45 (58.5%) | 2.34 (72.4%) | 2.35 (21.7%) |
| SRC(Homotopy) | 13.98 (65.0%) | 13.73 (37.5%) | 3.56 (60.0%) | 3.59 (17.3%) |
| GSRC[20] | 119.32 (93.0%) | 118.05 (79.0%) | 12.95 (66.2%) | 12.49 (75.5%) |
| RRC_$L_1$ | 8.70 (**100%**) | 8.62 (**97.5%**) | 2.06 (**94.2%**) | 2.04 (84.8%) |
| RRC_$L_2$ | **2.17** (99.5%) | **2.04** (96.5%) | **0.23** (91.5%) | **0.23 (86.0%)** |

## 4.6. Parameter discussion

In this section, we discuss the effect of parameter $\delta$ in RRC on the final recognition rate. As described below Eq. (14) and in Section 4.1, the parameter $\delta$ is a key parameter to distinguish inliers or outliers (if the residual's square of a pixel is larger than $\delta$, its weight will be less than 0.5; otherwise, its weight is bigger than 0.5). In our implementation, we use the parameter $\tau$ to estimate $\delta$, as described in Eq. (24). Hence, it is necessary to discuss the selection of $\tau$. Here we take the experiment with various level random pixel corruption (experimental settings are described in Section 4.3.1) as an example to discuss the selection of $\tau$ for RRC. Fig. 8 plots the recognition rates of RRC_$L_1$ versus different values of $\tau$ for 0%, 30%, 60%, and 90% pixel corruption. It can be seen that for moderate corruption (i.e., 0%~60%), RRC_$L_1$ could get very good performance (i.e., more than 95%) in a broad range of $\tau$. For all percentages of pixel corruption, the best performance could be achieved when $\tau$=0.5. Compared to CESR [33], whose kernel size is very sensitive to the corruption percentage (please refer to Section 5.7 of [33]), our proposed RRC method is easy to tune and is more robust to occlusion. Usually the domain of $\tau$ could be set as [0.5, 0.8]. It is reasonable because at least 50% samples should be trusted when there are large percent of outliers.

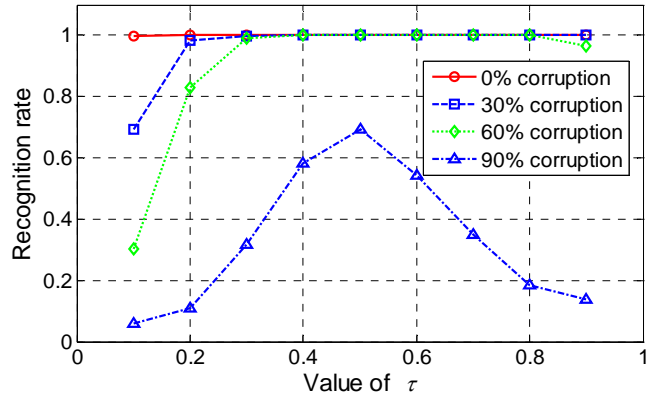

**Figure 8:** Recognition performance versus $\tau$ in estimating $\delta$ of RRC's weight function.

## 5. Conclusion

This paper presented a novel robust regularized coding (RRC) model and an associated effective iteratively reweighted regularized robust coding (IR$^3$C) algorithm for robust face recognition (FR). One important advantage of RRC is its robustness to various types of outliers (e.g., occlusion, corruption, expression, etc.) by seeking for an approximate MAP (maximum a posterior estimation) solution of the coding problem. By assigning adaptively and iteratively the weights to the pixels according to their coding residuals, the IR$^3$C algorithm could robustly identify the outliers and reduce their effects on the coding process. Meanwhile, we showed that the $l_2$-norm regularization is as powerful as $l_1$-norm regularization in RRC but the former has much lower computational cost. The proposed RRC methods were extensively evaluated on FR with different conditions, including variations of illumination, expression, occlusion, corruption, and face validation. The experimental results clearly demonstrated that RRC outperforms significantly previous state-of-the-art methods, such as SRC, CESR and GSRC. In particular, RRC with $l_2$-norm regularization could achieve very high recognition rate but with low computational cost, which makes it a very good candidate scheme for practical robust FR systems.